%% file: root.tex
\documentclass[letterpaper, 10 pt, conference]{ieeeconf}

\IEEEoverridecommandlockouts
\usepackage{amssymb}

\usepackage{amsthm}
\usepackage{amsmath}
\usepackage{multirow}
\usepackage{threeparttable}
\PassOptionsToPackage{hyphens}{url}\usepackage{hyperref}
\usepackage[dvipsnames]{xcolor}
\DeclareMathOperator*{\argmin}{arg\,min}

\DeclareUnicodeCharacter{0301}{}
\usepackage{graphicx}
\usepackage{caption}
\usepackage[caption=false,font=normalsize,labelfont=sf,textfont=sf]{subfig}

\begin{document}
%
\title{Direct and Sparse Deformable Tracking}
%
%
%

\author{José Lamarca,
        Juan J. Gómez Rodríguez,
        Juan D. Tardós, ~\IEEEmembership{Member,~IEEE,} and~J.M.M. Montiel,~\IEEEmembership{Member,~IEEE}
\thanks{This work was supported by the EU-H2020 grant 863146: ENDOMAPPER, the Spanish government grants PGC2018-096367-B-I00, DPI2017-91104-EXP and the MINECO scholarship BES-2016-078678, and by Aragón government grant DGA\_T45-17R.}
\thanks{The authors are with the Instituto de Investigaci\'on en Ingenier\'ia de Arag\'on (I3A), Universidad de Zaragoza, 
Mar\'ia de Luna 1, 50018 Zaragoza, Spain. E-mail: \{jlamarca,  jjgomez, tardos, josemari\}@unizar.es.} }

\markboth{Journal of \LaTeX\ Class Files,~Vol.~14, No.~8, August~2015}%
{Shell \MakeLowercase{\textit{et al.}}: Bare Demo of IEEEtran.cls for IEEE Journals}

\maketitle

\begin{abstract}
Deformable Monocular SLAM algorithms recover the localization of a camera in an unknown deformable environment. Current approaches use a template-based deformable tracking to recover the camera pose and the deformation of the map. These template-based methods use an underlying global deformation model. In this paper, we introduce a novel deformable camera tracking method with a local deformation model for each point. Each map point is defined as a single textured surfel that moves independently of the other map points. Thanks to a direct photometric error cost function, we can track the position and orientation of the surfel without an explicit global deformation model. In our experiments, we validate the proposed system and observe that our local deformation model estimates more accurately and robustly the targeted deformations of the map in both laboratory-controlled experiments and in-body scenarios undergoing non-isometric deformations, with changing topology or discontinuities.
\end{abstract}


%
\IEEEpeerreviewmaketitle

\input{sections/introduction}

\input{sections/related}

\input{sections/formulation_mt}
\input{sections/shape-from-template}
\input{sections/deformation_tracking}

\input{sections/experiments}
\input{sections/conclusions}



%



\bibliographystyle{IEEEtran}
\bibliography{IEEEabrv,bibl}
\end{document}

%% file: sections/introduction.tex
\section{Introduction}

VSLAM (Simultanenous Localization And Mapping from Visual sensors) is becoming a mature technology to navigate in human-made environments, being crucial for technologies like augmented reality and autonomous robot operation. Current state-of-the-art VSLAM algorithms \cite{campos2020orb,engel2017direct,zubizarreta2020direct} strongly rely on scene rigidity. As a consequence, they  perform poorly in deforming scenes, e.g. in  medical environments.

Since PTAM \cite{klein2007parallel}, VSLAM algorithms divide the computation in a tracking and a mapping concurrent threads. The tracking thread computes the camera position \textit{wrt.} the map at frame-rate. In parallel, the mapping thread recovers the structure of the scene with a higher computational cost from some selected frames, so-called keyframes. In the deformable case, both DefSLAM \cite{lamarca2019defslam} and SD-DefSLAM \cite{rodriguez2020sd} use a deformable mapping based on a Non-Rigid Structure-from-Motion (NRSfM) \cite{parashar2017isometric} to recover the structure of the scene at keyframe rate, and a deformable tracking \cite{lamarca2018camera} that estimates simultaneously the camera pose and the deformation of the map for every frame. 

The deformable tracking of these previous methods relies on the usage of a mesh that embeds the map points, and it recovers the most likely shape of the mesh according to a deformation model. This deformation model is global i.e. each map point is connected with their neighbours. This shows excellent performance in scenes with a single surface where all the points are indeed connected. However, when points are not connected, like in scenes with several surfaces, non-isometric surfaces, or with topological changes, the global model does not represent properly the deformation of the map, yielding low performance. 

In this paper, we propose a novel deformable tracking method that uses local deformation models to treat the map points as independent bodies. Our first contribution is to model the map as a sparse set of 3D moving textured surfels observed by a moving perspective camera. Each surfel is assumed to have independent rigid displacements from the other surfels around its position at rest. The formulation of the surfel is a first-order Taylor approximation of the map point. The main advantage of this approach is that any smooth surface, e.g. cylinders, planes, spheres or discontinuous surfaces, can be represented locally by a plane, independently of its topology.

\begin{figure}
    \centering
    \includegraphics[width=\columnwidth]{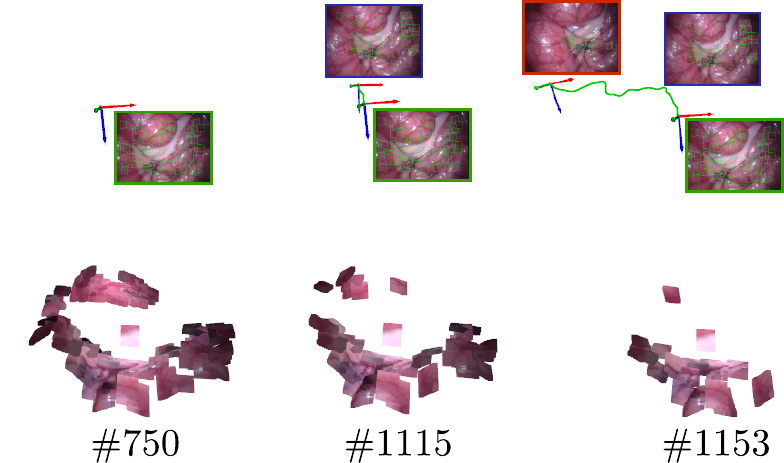}
    \caption{Direct and Sparse Deformable Tracking processing Hamlyn Dataset sequence 6, results  after frames \#750, \#1115 and \#1153. Bottom: The map composed of sparse surfels. Top: Camera Trajectory in green.}
    \vspace{-0.70cm}
    \label{fig:teaser}
\end{figure}

Our second contribution is to use a direct photometric error resulting from back-projecting the surfel texture. We jointly optimize the 3D position and orientation of the surface to minimize the direct photometric error. In contrast to previous approaches, in our proposed direct deformable tracking there is no hard data association, instead, the final matching is a byproduct of the photometric alignment. 
In our experimental section, we prove that our method can deal with discontinuous surfaces and topological changes, and achieves better performance than the tracking method used in current Deformable Monocular SLAM methods \cite{lamarca2019defslam,rodriguez2020sd}, obtaining longer tracks with better geometrical accuracy in medical sequences.

Next, in Sec.\ref{sec:related_work}, we discuss in detail the related works in non-rigid reconstruction and VSLAM. In Sec.\ref{sec:formulation}, we present our formulation for the surfel. In Sec.\ref{sec:sub_sft}, we develop our deformable tracking with fixed camera to prove the potential of surfel tracking adapting to different surfaces. In contrast to the previous methods, we propose a fully direct and sparse approach able to recompute the matches during the optimization. In Sec.\ref{sec:sub_dt}, we formulate a world-centric direct deformable tracking to estimate the pose of the camera based on a equilibrium regularizer. Finally, in the last Sec.\ref{sec:experiments_}, the results obtained show a considerable improvement \textit{wrt.} the previous deformable tracking methods both in terms of robustness and accuracy.


%% file: sections/related.tex
\section{Related work}
\label{sec:related_work}
Deformable SLAM problem consists in reconstructing a map whose shape is constantly deforming and recovering the camera trajectory \textit{wrt.} the reconstructed map.

The first deformable SLAM method proposed was DynamicFusion \cite{newcombe2015dynamicfusion}. This method proposes a pipeline where the entire shape of map was reconstructed from partial {RGB-D} observations from different positions. MISSLAM \cite{song2018mis} transferred this technique to medical scenarios by using stereo pairs. Concerning monocular SLAM, the lack of depth information significantly entangles the reconstruction problem. The first work to solve Deformable SLAM with monocular cameras was DefSLAM \cite{lamarca2019defslam}. Like other monocular SLAM systems \cite{campos2020orb,engel2017direct,zubizarreta2020direct}, DefSLAM is composed of two main threads: deformable tracking and mapping. These two components are based on the two main families of non-rigid monocular methods: Non-Rigid Structure-from-Motion for mapping, and template-based techniques for tracking.  

The first approaches of NRSfM were formulated using statistical models  \cite{bregler2000recovering,dai2014simple,akhter2011trajectory}. A low dimensional basis model is used to obtain the configuration of the 3D points for several images. The problem has been formulated with different regularizers, e.g. spatial \cite{dai2014simple,garg2013dense}, temporal \cite{akhter2011trajectory}, or spatio-temporal \cite{ gotardo2011kernel}. The main weakness of these methods is the assumption of orthographic camera model, not suitable for VSLAM due to the noticeable perspective effects in many targeted scenes where close-ups are dominant. Recent geometric methods have been proved to work with perspective cameras under the assumption of local isometry in the surface \cite{chhatkuli2014non,chhatkuli2016inextensible,parashar2017isometric,taylor2010non,vicente2012soft}. The method proposed in \cite{parashar2017isometric} was the base of the deformable mapping in \cite{lamarca2019defslam} due to its ability to naturally handle  occlusions and missing data . 

Template-based techniques recover the deformation of the scene from a single-image relying in a known textured surface and a deformation model. The 3D shape at rest of the textured surface is the so-called template. In the deformable SLAM approaches, the template is used to estimate the deformation of the map during tracking. The main difference between these methods is the representation of the surface and its deformation model. Among the analytic solutions, one of the most extended assumptions is that the surface is isometric. In other words, the geodesic distance between points in the surface is preserved during the entire sequence. Isometry for shape-from-template --SfT-- has been proven to be well-posed and to quickly evolve to stable and real-time analytical solutions \cite{bartoli2015shape,chhatkuli2017stable,collins2010locally}. On the other hand, energy-based methods \cite{salzmann2011linear} are numerical approaches that jointly minimize the shape deformation energy \textit{wrt.} the shape-at-rest and the reprojection error for the current image correspondences. These optimization methods are well suited to implement sequential data association with robust kernels to deal with outliers. 

The mentioned methods consider the camera static and usually reconstruct small objects that move in the camera field of view. The deformable tracking methods estimate the camera pose in addition to the deformation of the map. Usually, this is done by constraining the problem with boundary conditions  \cite{agudo2014good,lamarca2018camera}. The deformable tracking for deformable monocular SLAM \cite{lamarca2019defslam, rodriguez2020sd} was built on top of \cite{lamarca2018camera}. Template-based methods rely on a global model that connects all the map points and are prompt to fail when the map points are simply not connected or have a different relation.  

In this paper, we formulate the points of the surface separately as surfels -surface element- and jointly estimate its position for each frame and the position of the camera. One of the closest approach was the scene flow technique proposed in \cite{devernay2006multi}, that uses surfels to track some points of the scene, however they rely in a multi-camera setup, while we use a monocular camera. Using surfels, we can represent more general disconnected shapes of the scene and movements, and avoid the usage of a global deformation model. 

Piecewise methods are local techniques where the non-rigid object is a collection of pre-defined patches that move independently as rigid objects. The first work in using this strategy was \cite{varol2009template}, imposing a 3D global consistency in overlapping points. A relaxation to the
piecewise rigid constraint was given by \cite{fayad2010piecewise}, assuming each patch deforms with a quadratic physical model accounting for linear and bending deformations. All these methods required an initial patch segmentation and
the number of overlapping points, to this end \cite{russell2011energy} optimize the number of patches and overlapping through an energy-based optimization. In contrast, \cite{taylor2010non} constructs a triangular mesh, connecting all the points, and considering each triangle as being locally rigid, being able to deal with topological changes. Our method belongs to this family of methods, but in contrast, we do not assume that the points are overlapping. 

Our previous work \cite{rodriguez2020sd} is a semi-direct method that replaces the feature-based tracking of \cite{lamarca2019defslam} with a multiscale Lucas-Kanade tracker, resulting in an improvement of the track lengths and reconstruction accuracy. In this work, we take advantage of direct photometric error to recover the 3D relative position of the surface points. Direct methods use the photometric error and have been proven extremely accurate in the rigid SLAM case \cite{engel2017direct,zubizarreta2020direct} and other NRSfM works like \cite{yu2015direct}.

%% file: sections/formulation_mt.tex
\section{Formulation}
\label{sec:formulation}
\begin{figure}
    \vspace{2mm}
    \centering
    \includegraphics[width=0.9\columnwidth]{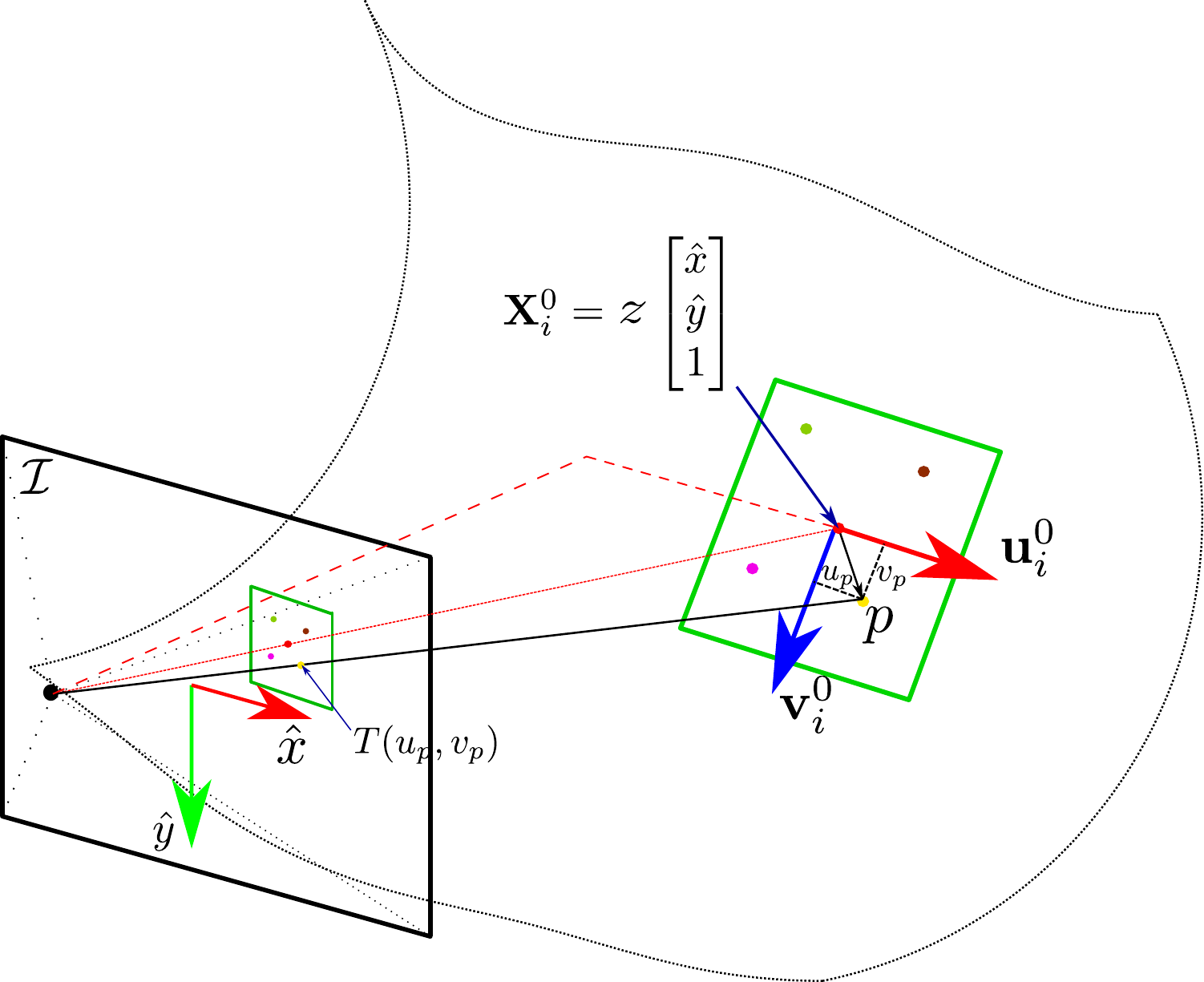}
    \caption{Parametrization of a surfel in the initial image. Coordinates of the surface $u$ and $v$ correspond to the normalized coordinates in the image $\hat{x}$ and $\hat{y}$. We obtain $z$ from the depth image, and we estimate the tangent space vectors ${\mathbf{u}}_i^t$ and ${\mathbf{v}}_i^t$ as the directional derivatives in the image coordinates $\mathcal{I}$.}
    
    \vspace{-0.5cm}
    \label{fig:SurfelBA}
\end{figure}

This section is devoted to formalize the parametrization of a surfel and  the photometric equations describing its observation by  a projective camera.

\subsection{Notation} Bold letters represent vectors or matrices ($\mathbf{X}$). Scalars will be represented by light lowercase letters ($t$), image brightness functions by light uppercase letters ($I$). Superindex $t$ denotes the frame in which the estimation is done. Subindex $i$ identifies the surfel. Subindex $p$ refers to pixel coordinates in reference local to the surfel. To simplify the index notation all the scene points coordinates are in the world reference. Camera poses are represented as transformation matrices $\mathbf{T}_{cw} \in SE(3)$, transforming the coordinates of point from the world frame into the camera frame. 

\subsection{Surfel parametrization}
Assuming a continue and derivable $C^{1}$ surface, a point $\mathbf{X}_i^t$ is represented by a surfel $\mathbf{S}_i^t$ contained in the tangent space of the surface at the point. Thus, a generic 3D point $p$ belonging to the surfel can be parametrized using two local coordinates $u_p$ and $v_p$ around $\mathbf{X}_i^t$:

\begin{equation}
    \mathbf{S}_i^t(u_p,v_p) = \mathbf{X}_i^t + \mathbf{J}_i^t \begin{bmatrix}
    u_p \\ v_p 
    \end{bmatrix}
    \label{eq:patch3d_transform}
\end{equation}
\begin{eqnarray}
    \mathbf{J}_i^t &=& \begin{bmatrix}
    {\mathbf{u}_i^t} & {\mathbf{v}_i^t}
    \end{bmatrix} \in \mathbb{R}^{3\times 2}
\end{eqnarray}

\noindent where  $\mathbf{J}_i^t$ is the so-called Jacobian matrix whose columns are a pair of vectors forming a base of the tangent space. As described in Eq. (\ref{eq:patch3d_movement_sft}), $\mathbf{X}_i^t$ and $\mathbf{J}_i^t$ are defined for each frame in terms of the corresponding values at $t=0$, $\mathbf{X}_i^0$ and $\mathbf{J}_i^0$, whose initialization from the first image is described next.

%
\subsection{Surfel initialization}
We assume the scene surface is defined by means of the depth function: $z(\hat{x},\hat{y}):\mathbb{R}^{2}\to \mathbb{R}$ in terms of the normalized retina coordinates $\hat{x},\hat{y}$. This depth function can be provided by a depth sensor (RGB-D camera or stereo rig). Then, $\mathbf{X}_i^0$ and $\mathbf{J}_i^0$ are estimated as:

\begin{equation}
    \mathbf{X}_i^0 = z(\hat{x},\hat{y}) \begin{bmatrix} \hat{x}\\\hat{y}\\1
    \end{bmatrix} 
    \label{eq:position}
\end{equation} and
\begin{equation}
    \mathbf{J}_i^0 =  \begin{bmatrix} 
    z + \hat{x} \frac{\partial z}{\partial \hat{x}} & \hat{x} \frac{\partial z}{\partial \hat{y}}\\
    \hat{y} \frac{\partial z}{\partial \hat{x}} & z + \hat{y} \frac{\partial z}{\partial \hat{y}}\\
    \frac{\partial z}{\partial \hat{x}} & \frac{\partial z}{\partial \hat{y}}
    \end{bmatrix}
    \label{eq:Jacobian}
\end{equation}

For the experiments, we initialize surfels in the interest points extracted with Shi-Tomasi \cite{shi1994good}.

\subsection{Photometric error}
We denote the projection function as $\pi(\cdot):\mathbb{R}^{3}\to \mathbb{R}^{2}$. For our experiments, we use the pinhole camera model. Note that this can be easily substituted by any other camera model.


We optimize the difference between the intensities of points in the surfel and the intensities in their reprojections in the current image:

\begin{equation}
    \mathcal{P}^t_i =  \sum_{p} \left(\alpha_i^t I_t\left(\pi\left(\mathbf{T}_{cw} \mathbf{S}_i^t(u_p,v_p)\right)\right)+ \beta_i^t -T(u_p,v_p)\right)^2
    \label{eq:photometric_error_illu}
\end{equation}

\noindent where $\mathbf{T}_{cw}$ is the pose of the camera with respect to the world. $\mathbf{S}_i^t(u_p,v_p)$ is in the world reference. 
We compensate the illumination changes by means of a gain ($\alpha_i^t$) and a bias ($\beta_i^t$ ) per surfel and per image. 
That allows us to synthesize the deformed surfel into the image, thus our error function takes into account the local deformation.

We define a symmetric uniform grid in the surfel local coordinates that is reprojected into the inital image to extract the surfel texture $T(u_p,v_p)$  parameterized by $u_p$ and $v_p$.

%% file: sections/shape-from-template.tex
\section{Direct and Sparse camera tracking with static camera}
\label{sec:sub_sft}
Let's assume in this section that the camera is fixed and the initial values of the textured surfels are given in advance, and we want to estimate the deformation for each incoming image. With our formulation, the initial surfel is defined by its initial position $\mathbf{X}_i^0$, its Jacobian $\mathbf{J}_i^0$ and its texture $T(u_p,v_p)$. 

The geometrical transformation of the surfel is expressed as:
\begin{equation}
    \mathbf{S}_i^t(u_p,v_p) =  (\mathbf{X}_i^0+ \mathbf{t}_i^t) + \mathbf{R}_i^t \mathbf{J}_i^0 \mathbf{F}^t_i \begin{bmatrix}
    u_p \\ v_p
    \end{bmatrix}
    \label{eq:patch3d_movement_sft}
\end{equation}
where $\mathbf{t}_i^t \in \mathbb{R}^3$ is the translation of the surfel, $\mathbf{R}_i^t \in \mathbf{SO}(3)$ is the rotation of the surfel modeled by the 3 parameters of its Lie algebra, and $\mathbf{F}^t_i \in \mathbb{R}^{2\times2}$ is a symmetric matrix that represent the deformation tensor. Its diagonal components represent the stretching of the tangent vectors, and its off-diagonal element models the angle change between these vectors, i.e. the shearing. 

\input{tables/tableDefModels}
As seen in Table \ref{tab:Constrain}, the most restrictive local deformation is isometric. This constraint is equivalent to a rigid movement of the surfel. When the surfel deformation is not bounded the first ambiguity arises:

\textbf{Growing map ambiguity.} The depth component of the translation of the surfel and the surfel size can be coupled in such a way that changing its depth and size produces the same image. 

\begin{proof}
We define an $\mu$ factor that transforms the position and the deformation of the surfel as:
\begin{eqnarray}
    (\mathbf{X}_i^0 + \mathbf{t}_i^t) &=& \mu \mathbf{X}_i^0\\
    \mathbf{F}^t_i &=& \begin{bmatrix}
    \mu & 0 \\ 0 & \mu
    \end{bmatrix}\\
    \hat{\mathbf{S}}_i^t(u_p,v_p) &=&\mu \mathbf{X}_i^0 + \mu \mathbf{R}_i^t \mathbf{J}_i^0 \begin{bmatrix}
    u_p \\ v_p
    \end{bmatrix} = \mu \mathbf{S}_i^t(u_p,v_p)
    \label{eq:proportion}
\end{eqnarray}

Under perspective projection any surfel $\hat{\mathbf{S}}_i^t(u_p,v_p)$ multiplied by $\mu$ produces the same image $\pi(\mathbf{S}_i^t(u_p,v_p)) = \pi(\hat{\mathbf{S}}_i^t(u_p,v_p))$.
\end{proof}

To solve this ambiguity, we impose local isometry within the surfel. Isometry is a distance preserving transformation. We propose two alternatives to code the isometry, as a hard constraint or as a soft constraint.

Isometry as hard constraint implies the transformation of the surfel only as a rigid body motion, in other words, the deformation matrix, $\mathbf{F}^t_i = \mathbb{I}_2$. The motion is defined by 6 parameters (3 for translation + 3 for rotation). 
\begin{equation} \label{eq:hard_constrain}
\begin{split}
\mathbf{J}_i^t &= \mathbf{R}_i^t \mathbf{J}_i^0 \;\;\; ; 
  \;\;\; \mathbf{R}_i^t \in \mathbf{SO}(3) 
\end{split}
\end{equation}

Thus, our cost function is defined only by the photometric error (Eq.~(\ref{eq:photometric_error_illu})):
\begin{equation}
    \mathbf{t}_i^t , \mathbf{R}_i^t = \underset{\mathbf{t}_i^t , \mathbf{R}_i^t, \alpha_i, \beta_i}{\argmin} \mathcal{P}^t_i 
    \label{eq:hard_constrain_error_function}
\end{equation}

In the case of a soft constrain we penalize the stretching and shearing of the surfel. It is formulated through the tangent plane $\mathbf{J}_i^t$. We define a deformation energy quadratic error as:
\begin{equation}
    \mathcal{I}^t_i = \left\|\mathbf{F}_i^t - \mathbb{I}_2 \right\|_2^2
    \label{eq:soft_constrain}
\end{equation}

The soft constraint is modeled by means of the deformation energy coded by 3 additional parameters defining the symmetric matrix $\mathbf{F}^t_i$. In other words, the surfel can stretch and shear if it explains better the image, but it tends to stay as close as possible to its original shape. $\mathcal{I}^t_i$ is a scalar that penalises the shearing and stretching. 

Finally, the optimization is a combination of the forward-compositional photometric error and the deformation energy. The deformation energy regularization is weighted by a constant $\omega_{\mathcal{I}}$,
\vspace{-0.1cm}
\begin{equation}
    \mathbf{t}_i^t , \mathbf{R}_i^t, \mathbf{F}_i^t = \underset{\mathbf{t}_i^t , \mathbf{R}_i^t, \mathbf{F}_i^t ,\alpha_i, \beta_i}{\argmin} \mathcal{P}^t_i + \omega_{\mathcal{I}}\mathcal{I}^t_i 
    \label{eq:soft_constrain_error_function}
\end{equation}
\vspace{-0.1cm}
All the errors considered in (\ref{eq:hard_constrain_error_function}, \ref{eq:soft_constrain_error_function}) are quadratic, so it can be solved as a non-linear least-squares problem. We propose  Levenberg–Marquardt (LM) optimization \cite{nocedal2006numerical}. The LM algorithm is a trust-region method that combines a Gauss-Newton and steepest descend. The step control is defined through the damping factor $\lambda$ that weights both methods, $\lambda$ also allows to control the step size. The Hessian is approximated as $H \approx  J^\top J$. 

During the optimization, the data association between the images is changed boosting the accuracy, however the convergence basin of the photometric optimization is small. We propose an strict step size control to avoid leaving the convergence basin. We confine the step to an ellipsoidal trust region defined by the diagonal matrix $D_{w} = \text{diag}(H)$. We apply a step policy where $\lambda$ is limited to be $\geq 1$ during the first steps to avoid long steps when far from the minimum. In a subsequent stage $\lambda$ is allowed to be reduced in order to benefit from the Gauss-Newton quadratic convergence. 

A singular values analysis of the Hessian matrix points out that this matrix is ill-conditioned, i.e. the ratio between the smallest and biggest singular values is $\ll1$. This reflects a different scaling in translation, rotation and deformation parameters. Thus, we propose to use a diagonal scaling preconditioner matrix $D_s(i,i) = \frac{1}{\sqrt{s_i}}$ to avoid numerical issues, being $s_i$ the diagonal values of $(H+\lambda D_w)$. At each iteration the $\mathbf{\delta x}$ is then estimated as:

\begin{eqnarray}
   D_s\left(H+\lambda D_w\right)D_s \mathbf{\delta x^*} &=& -D_s J^\top \mathbf{r} \\
   \mathbf{\delta x} &=& D_s \mathbf{\delta x^*}
\end{eqnarray}

In addition, to avoid mismatches due to discontinuities and light reflections, we saturate the photometric error. We also carry out a multi-scale optimization to increase the convergence basin observing that in case of temporal discontinuities or fast movements the algorithm becomes much more robust. 

We detect the outlier surfels using a threshold in the Zero Normalized Cross correlation (ZNCC) between the texture of the surfel and the texture of its reprojection because it is illumination invariant. If the ZNCC drops under a threshold the surfel can be assumed as badly tracked and the corresponding observation is marked as an outlier.

The algorithm complexity is linear in the number of points since each new point would suppose a new optimization and cubic in the number of pixels per surfel since the Jacobian block of the surfel is dense and it increases one row per new pixel included.


%% file: tables/tableDefModels.tex
\begin{table}[]
\centering
\caption{Deformation tensor $F^t_i$ for different local deformation models.}
\begin{tabular}{l|l|l|l|l|}
\cline{2-5}
                               & Isometry & Conformal & Equireal & General \\ \hline
\multicolumn{1}{|l|}{$F^t_i$} &    $\mathbb{I}_2$      &     $s \mathbb{I}_2$      & $\begin{bmatrix}
    \alpha & \beta \\ \beta & \frac{1+\beta}{\alpha}
    \end{bmatrix}$&  $\begin{bmatrix}
    \alpha & \beta \\ \beta & \gamma
    \end{bmatrix}    $   \\ \hline
\multicolumn{1}{|l|}{Variables} &    -      &     $s$      & $\alpha,\beta$ &$\alpha,\beta,\gamma$         \\ \hline
\end{tabular}
\label{tab:Constrain}
\vspace{-0.5cm}
\end{table}

%% file: sections/deformation_tracking.tex
\section{Direct and Sparse deformable tracking}
\label{sec:sub_dt}
Deformable tracking algorithm takes as input the textured surfels and the initial camera pose. Then, it estimates the deformation of the map and the camera pose \textit{wrt.} the map.

\textbf{Floating map ambiguity.} Surfel position and camera pose are coupled and can be varied producing the same projection of the pixel in the image.

\begin{proof}
Eq. (\ref{eq:patch3d_transform}) can be rewritten as:
\begin{equation}
    \mathbf{S}_i^t(u_p,v_p) = \begin{bmatrix}
        \mathbf{R}_i^t & \mathbf{t}_i^t \\ 
        \mathbf{0} & 1
    \end{bmatrix}
    \begin{bmatrix}
        \mathbf{J}_i^0 \mathbf{F}^t_i\begin{bmatrix}
    u_p \\ v_p
    \end{bmatrix} & \mathbf{X}_i^0 \\
        \mathbf{0} & 1
    \end{bmatrix} 
\end{equation}
where we can define the rigid movement of the surfel as $\mathbf{T}_{iw}^t \in \mathbf{SE}(3)$:
\begin{equation}
    \mathbf{T}_{iw}^t = \begin{bmatrix}
        \mathbf{R}_i^t & \mathbf{t}_i^t \\ 
        \mathbf{0} & 1
    \end{bmatrix}
\end{equation}

The camera pose $\mathbf{T}_{cw}$ and the transformation of a surfel $\mathbf{T}_{iw}^t$ are coupled and can be interpreted as a arbitrary movements of the camera or as a movement of the surfel. 
\begin{eqnarray}
    \left[\mathbf{S}_i^t(u_p,v_p)\right]_c &=& \mathbf{T}_{cw} \mathbf{T}_{iw}^t 
    \begin{bmatrix}
        \mathbf{J}_i^0 \mathbf{F}^t_i\begin{bmatrix}
    u_p \\ v_p
    \end{bmatrix} & \mathbf{X}_i^0 \\
        \mathbf{0} & 1
    \end{bmatrix} \\
\hat{\mathbf{T}}_{cw} &=& \mathbf{T}_{cw} \mathbf{T}_{iw}^t
= \mathbf{T}_{cw}^* \mathbf{T}_{iw}^{t^*}
\end{eqnarray}
\end{proof}

To avoid the ambiguity, we propose to soft-constrain each surfel position around an equilibrium position $\mathbf{X}_{e_i}^t$ with the regularizer $\mathcal{E}^t_i$: 
\begin{equation}
    \mathcal{E}^t_i = \left(\mathbf{X}_i^t - \mathbf{X}_{i}^0\right)^\top \Sigma^{-1}_{i} \left(\mathbf{X}_i^t - \mathbf{X}_{i}^0\right)
    \label{eq:temporal_consistency}
\end{equation}

This position gives a reference for the camera estimation. We can understand the camera movement in our approach as the global rigid movement, and the deformation of the surfels as movements around that equilibrium. $\Sigma_i$ is the covariance that the surfels can reach in its movement.

If the trajectory of the points along the sequence is known in advance, the equilibrium point can be estimated as their average position and its covariance. In the case that the position and covariance are unknown, we approximated as it is around the original position and select a heuristic covariance with the expected movement. Lower covariances lead to more rigid interpretation.

Similarly to Sec.\,\ref{sec:sub_sft}, it is possible code the isometry as a hard or as a soft constraint. The  optimization for the hard constraint case is:
\begin{equation}
    \mathbf{X}_i^t, \mathbf{J}_i^t , \mathbf{T}_{cw} = \underset{\mathbf{X}_i^t , \mathbf{J}_i^t, \alpha, \beta, \mathbf{T}_{cw}}{\argmin} \sum_{i\in \mathcal{X}} \mathcal{P}^t_i + \omega_{\mathcal{E}}\mathcal{E}^t_i 
    \label{eq:dt_optimization}
\end{equation}

The movement of the camera is defined by using Lie algebra of $\mathbf{SE}(3)$. We linearize in the solution for each step and update the pose each step as:
\begin{equation}
\hat{\mathbf{T}}_{cw} = \text{exp}(\zeta) \mathbf{T}_{cw}
\label{eq:lie_algebra}
\end{equation}

The optimization is done by using Levenger-Marquard. We again need to scale the parameters through $D_s$ and control the step with $\lambda D_w$.

%% file: sections/experiments.tex
\section{Experiments}
\label{sec:experiments_}
We evaluate the performance of the two proposed methods: Sparse Deformable Tracking with and without static camera, in rigid and deformable scenarios. We use sequences of laboratory-controlled scenarios from CVLab \cite{varol2012constrained} and sequences of intracorporeal scenes selected from the Hamlyn Dataset \cite{Mountney2010ThreeDimensionalTD}. A video with the results is provided as supplementary material\footnote{\url{https://drive.google.com/file/d/1XJFbLsp_76eGqDisJj8Sjcljaf3F1c94/view?usp=sharing}}. 

\subsection{Tuning}
\subsubsection{\textbf{Surfel size}}
Our primary assumption is that any surface can be locally approximated by the tangent plane. The accuracy of the approximation deceases with the distance to the centre of the surfel, hence it decreases with the surfel size. In contrast, bigger surfels allow more accurate estimates of the surfel geometry. Thanks to the saturation policy that we apply, we have noticed that even big surfels can accurately estimate the surfel geometry. We chose a surfel size of $\approx23$ pixels experimentally. In experiments in the Kinect paper dataset, we have observed that the error is reduced for bigger surfels, even if they do not fully accomplish the planarity assumption. Too big surfels lead to problems with spatial discontinuities in the scene. 

\subsubsection{\textbf{Multi-scale}}
\label{sec:sub_multiscale}
The convergence basin of the photometric methods is around one pixel, using multi-scale increases  it to more that one pixel in the finest scale. We use the solution of a coarse scale as the initial guess of the next finer scale. In the kinect paper and T-shirt datasets, we found several missing frames. That precludes the convergence for many surfels if only the finest scale is used. Using 3 scales, the algorithm converges despite the missing frames. 

\subsubsection{\textbf{Outlier rejection}}
\label{sec:sub_outlier}

We evaluate the ZNCC method to classify inliers as points that have converged correctly in the optimization. Positives are inliers and negatives are outliers. The ground truth of the correct tracks are classified through a threshold in the RMSE \textit{wrt.} the ground truth surfel trajectory. We show the ROC curve in Fig.\,\ref{fig:comparison_overview} right \textit{wrt.} varying the ZNCC threshold. Ideally, the more up to the left the curve is, the better the classifier is. We finally select a value for the ZNCC of 0.95 for the experiments in the Kinect dataset, and 0.85 for the Hamlyn dataset. 


\subsubsection{\textbf{Soft vs. Hard isometric constraint}}
In Sec.\ref{sec:formulation} we have discussed two ways of constraining the deformation of the surfel. We have validated them in different sequences. We have observed that a soft constrained deformation model does not improve the accuracy of the system. Isometry seems to be a good local approach for the local deformation of the surfels, and thanks to treating the points individually we can recover very different global deformations. For instance, in sequence 21 from Hamlyn Dataset we can cover non-isometric global deformations, or in the kinect Paper Dataset, we can track multiple objects, because we are treating each point individually. We use the isometry as a hard constrain (Eq. \ref{eq:hard_constrain_error_function} and Eq. \ref{eq:dt_optimization}) in all the rest of experiments.

\subsection{Deformable Tracking with Static Camera}
\begin{figure}
    \centering
    \includegraphics[width=0.490\columnwidth]{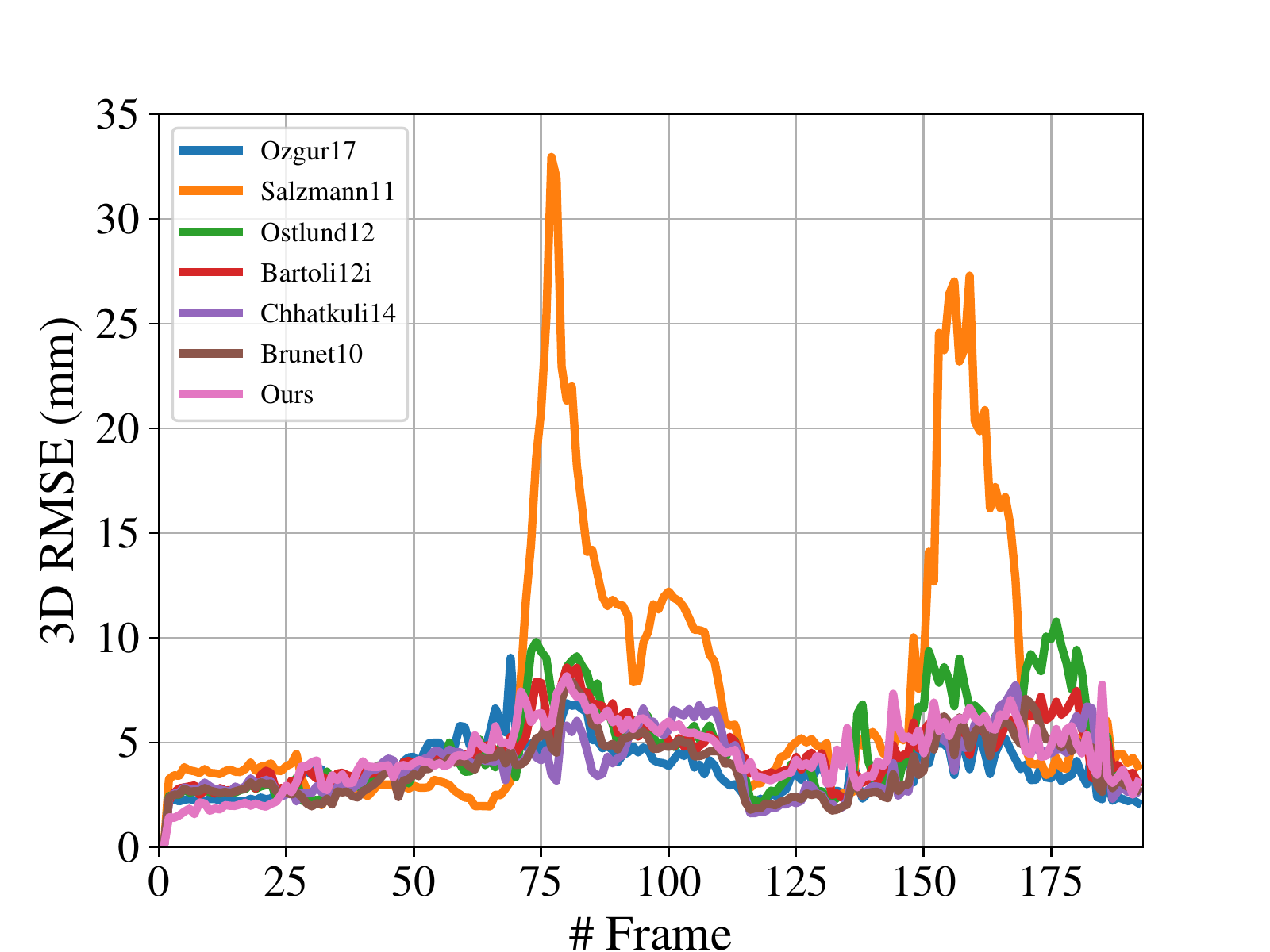}
    \includegraphics[width=0.490\columnwidth]{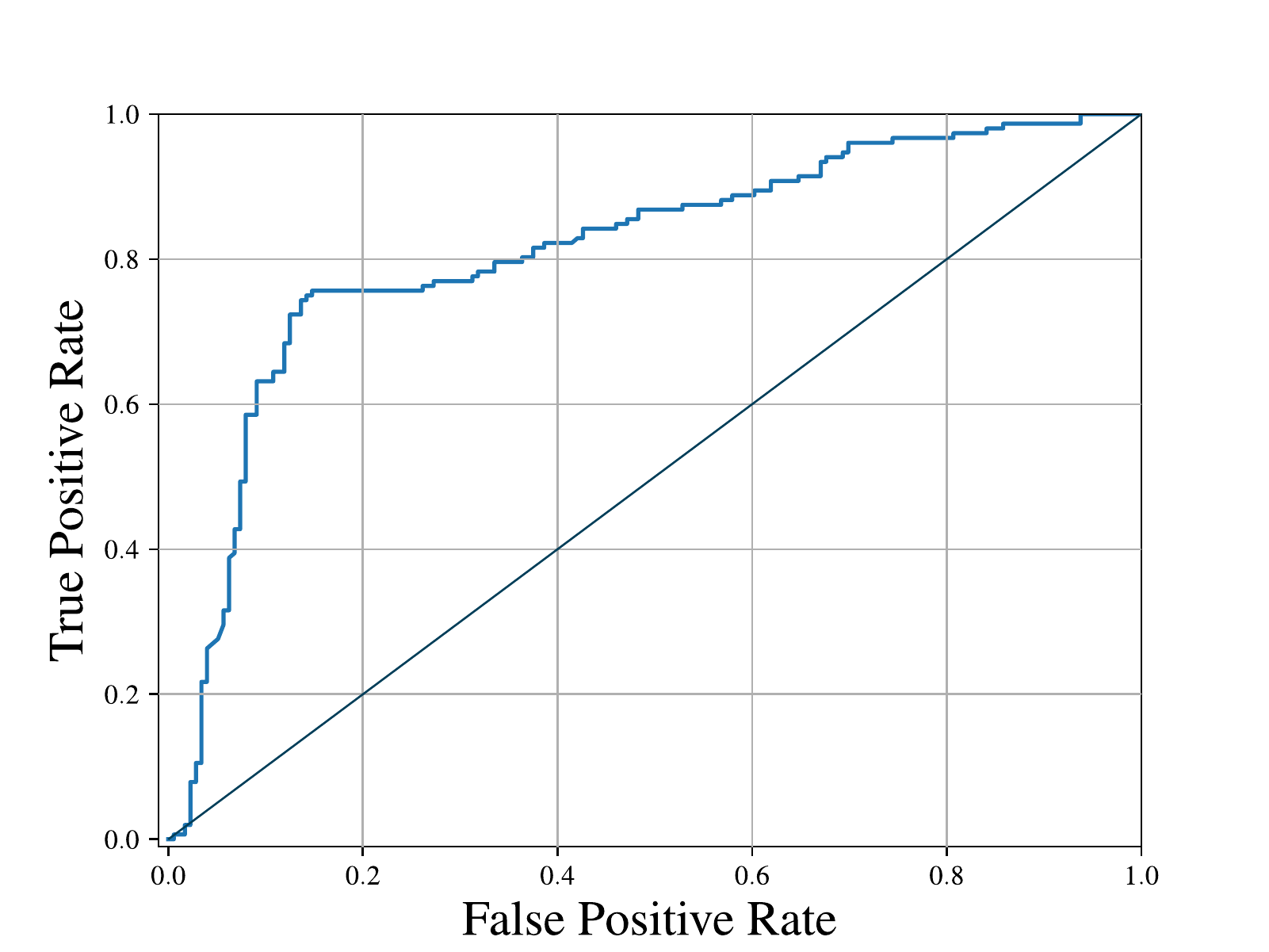}
    \caption{Left: Comparison in kinect Paper dataset from CVLab. Our method tracks individual surfels with similar accuracy than template-based methods that assume surface continuity. Right: Outlier detection, ROC curve \textit{wrt.} the ZNCC threshold}
    \vspace{-0.6cm}
    \label{fig:comparison_overview}
\end{figure}
\begin{figure*}
    \centering
    \includegraphics[trim= 0 1cm 0 0,clip,width=\textwidth]{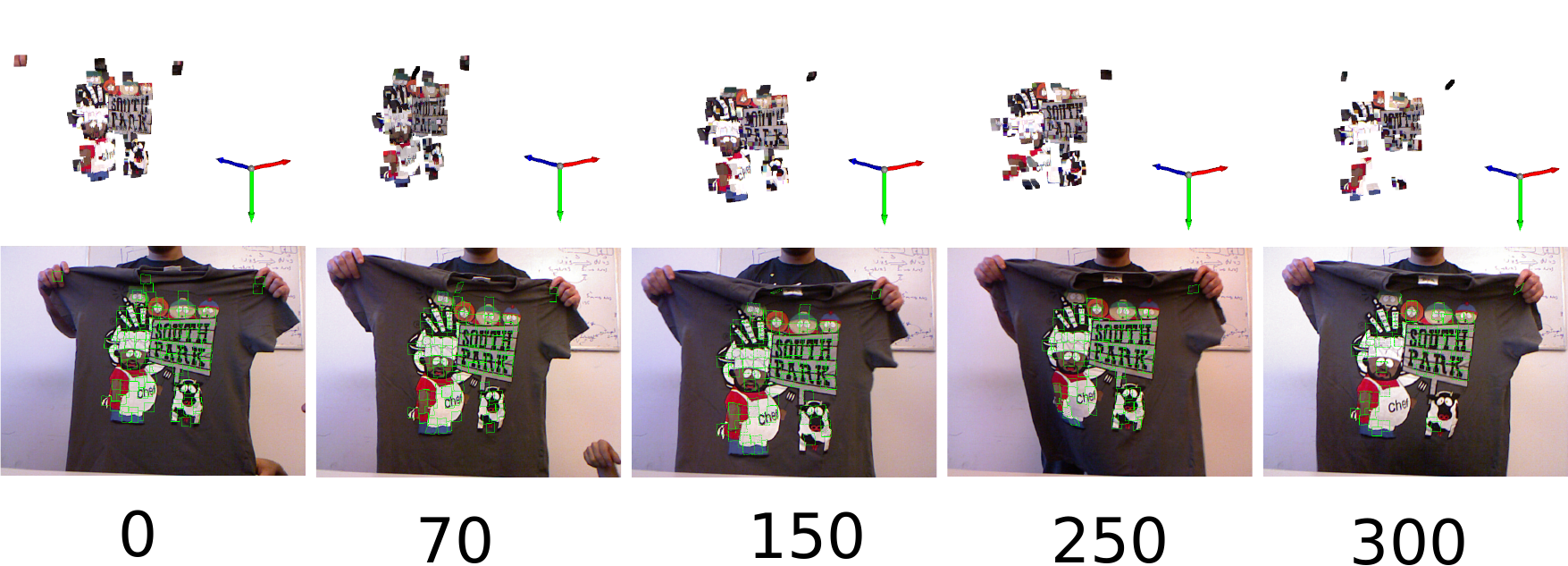}
    \includegraphics[trim= 0 1cm 0 0,clip,width=\textwidth]{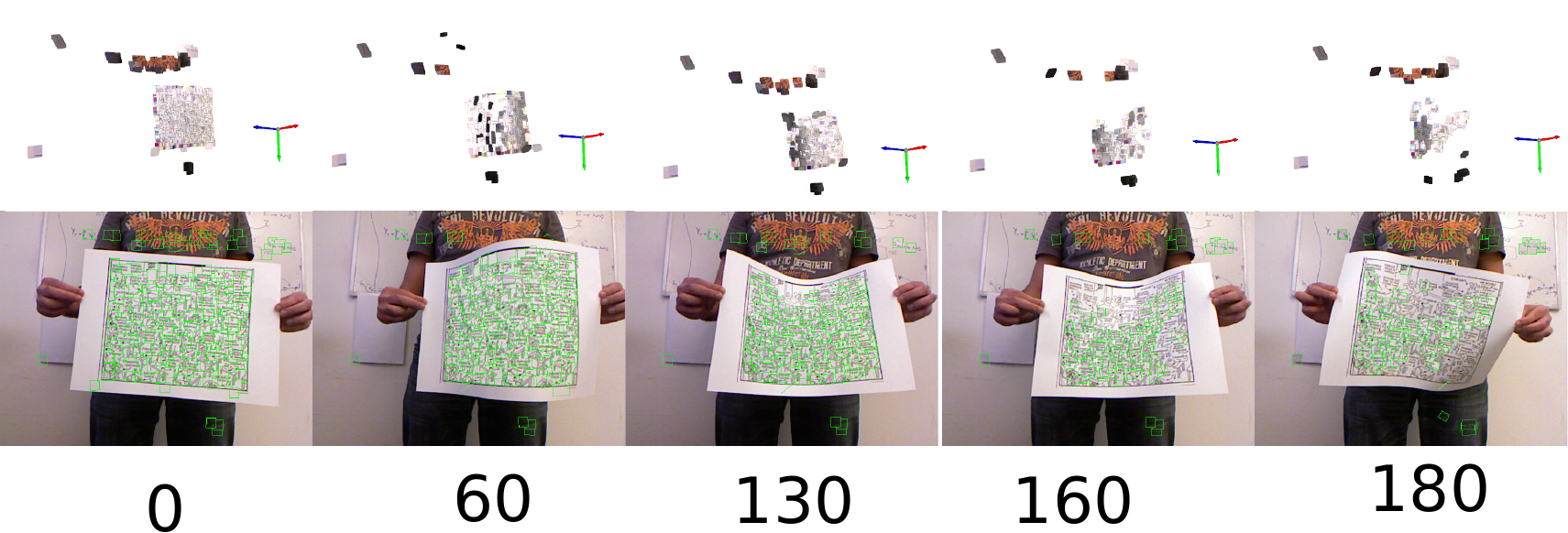}
    \includegraphics[trim= 0 1cm 0 0,clip,width=\textwidth]{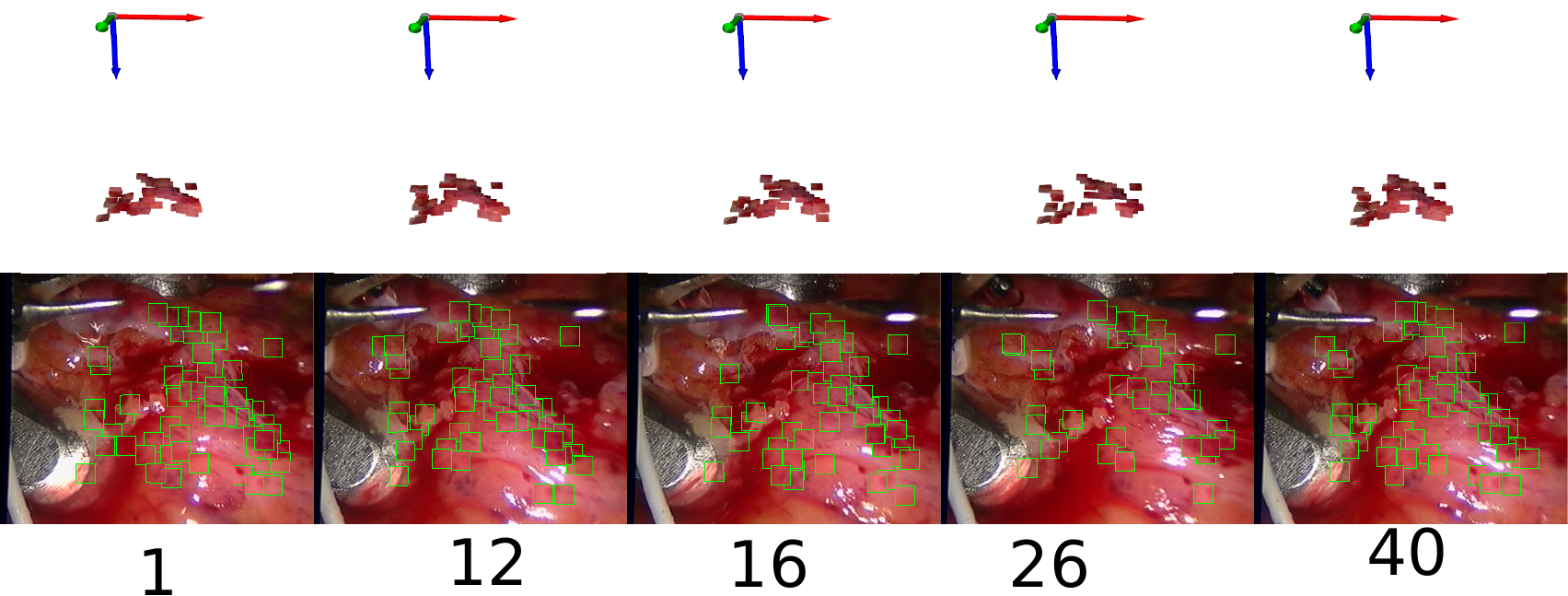}
    \includegraphics[trim= 0 1cm 0 0,clip,width=\textwidth]{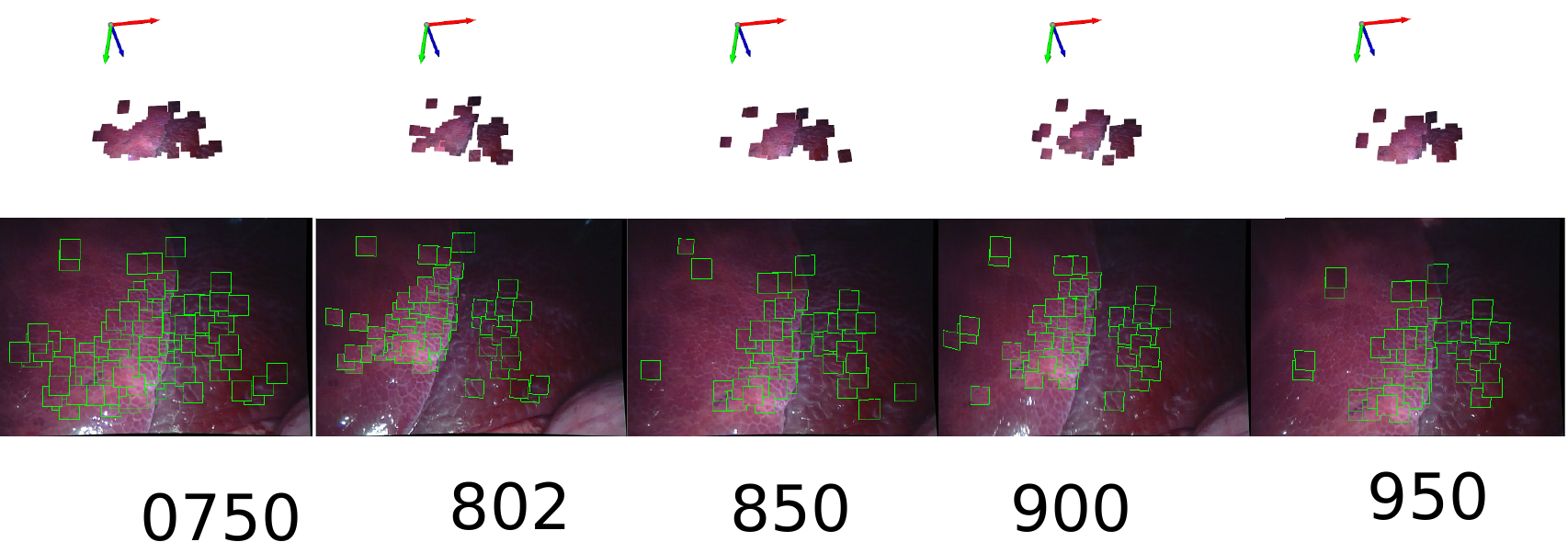}
    \caption{Deformable tracking results, two rows per sequence, first the 3D reconstruction, then the RGB frames. Even if the surfels are estimated independently the entire reconstruction displays  an homogeneous consistency. (See entire sequences in supplementary material). \\ \textbf{1st-2nd rows, kinect T-Shirt dataset}, frames \# 0,\# 70,\# 150,\# 250 and \#300. 
    \textbf{3rd-4th rows kinect Paper dataset}, frames \# 0,\# 70,\# 130,\# 160 and \#180. 
    \textbf{5th-6th rows Hamlyn 4 (Heart sequence)}, frames \# 0,\# 12,\# 16,\# 26 and \#40. 
    \textbf{7th-8th Hamlyn 21 (Liver sequence)}, frames \# 750,\# 800,\# 850,\# 900 and \#950.}
    \label{fig:results_sft}
\end{figure*}

In this section, we analyse the performance of deformable tracking with static camera in real sequences. The CVLab's, T-shirt and paper dataset were recorded with a Kinect RGB-D camera. We also test intracorporeal sequences from the Hamlyn Dataset, in this case with stereo camera. We use the first depth image to initialise the surfels, i.e. the position, Jacobian and texture of each surfel. Notice that our system is monocular, hence we only process the gray images obtained with the RGB-D camera or with the left camera. 


We compare our method against some reference shape-from-template (SfT) methods \cite{ozgur2017particle,salzmann2011linear,ostlund2012laplacian,bartoli12,Chhatkuli_2014_CVPR,brunet2010monocular} in the Kinect paper dataset from CVLab. This sequence consists in a paper deformed isometrically. This sequence has two main challenges for our method: illumination changes and temporal discontinuities (missing frames) in frames \#70, \#120, \#130 and \#150. In contrast to the other methods, ours is the only one using photometric error. 

As shown in Fig. \ref{fig:comparison_overview}, the first notable result is that our method can track individual surfels with similar accuracy to methods that assume smoothness in the surface. We also noticed that optimization-based methods \cite{salzmann2011linear,ostlund2012laplacian} get worse results than the rest. These methods assume sequential images and the missing frames break this assumption worsening the results. In our case, something similar happens, but thanks to the multi-scale configuration the convergence gets substantially improved. We also conclude that the local compensation of the illumination presented in Eq.\,\ref{eq:photometric_error_illu} is crucial to track a higher number of surfels.

We have seen that assuming smooth surfaces improves the results in the paper area. However, this precludes the usage of this method in discontinuous surfaces. In contrast, as we have not assumed any regularizer between the individual surfels, we can track surfels not only on the paper area, but also on the person's T-Shirt and on the white board\, (See Fig.\,\ref{fig:results_sft}). Discontinuities raise other challenges like occlusions that are successfully managed with the saturation of the photometric error. 

\subsection{Direct and Sparse Deformable Tracking}
In this section, we analyse the deformable tracking, and we compare the advantage of a world-centric (DSDT) approach (Sec. \ref{sec:sub_dt}) where the camera can move versus a camera-centric approach with static camera (DSDT-SC) (Sec. \ref{sec:sub_sft}).

\begin{figure}
    \vspace{2mm}
    \centering
    \includegraphics[width=\columnwidth]{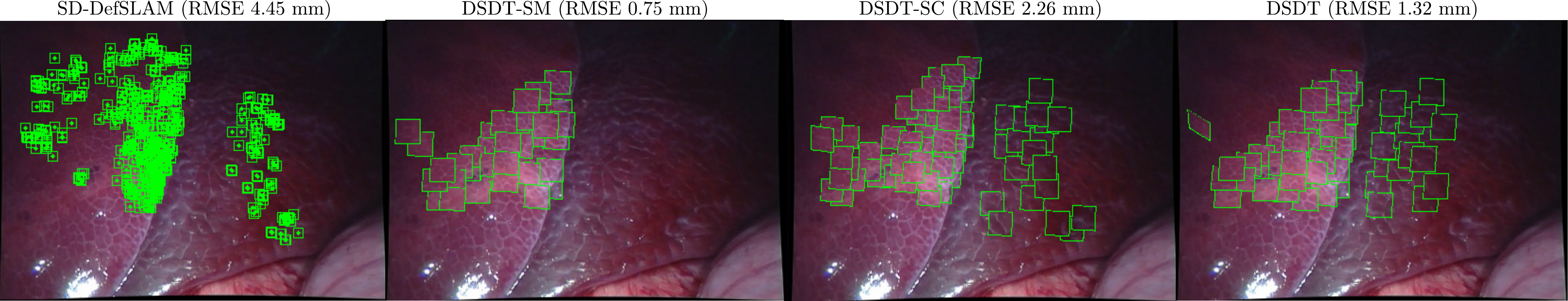}
    \caption{Non-isometric deformation results. Left to right: SD-DefSLAM, DSDT w. static map, w. static camera and DSDT.}
    \label{fig:niso}
    \vspace{-0.7cm}
\end{figure}
\input{tables/tabledt}

We compare both methods against our tracking with a static (i.e. rigid) map (DSDT-SM), the deformable tracking from SD-DefSLAM \cite{rodriguez2020sd} and the tracking of ORBSLAM \cite{mur2017orb} in the sequences 6, 20 and 21 from Hamlyn dataset \cite{Mountney2010ThreeDimensionalTD}. All the methods are initialized with a stereo pair in the same frame and no mapping is allowed, i.e., tracking the initial map without refining or extending it.

Dataset 6 (from frame $\#$50) is an abdominal exploration where the scene remains almost rigid. It has a planar topology in the area where the camera closes up, and a small discontinuity due to a nerve. The texture is minimal except for the veins. The deformable tracking can process 300 frames before tracking loss with an RMS error close to 3mm. On the other hand, DSDT-SC can process a similar number of frames but with a much bigger error. We conclude that the regularizer added in the deformation tracking gives hints to the optimizer yielding better performance. SD-DefSLAM processes a few frames less with similar error, however it only focuses on the planar area.

Dataset 20 (from frame $\#$750) is another abdominal exploration, but in this case the scene contains some global near-isometric deformation keeping a similar shape. Deformable tracking can track 500 frames from the initialization frame, in contrast to the camera-centric approach DSDT-SC that only process 350 frames. Again thanks to the movement of the camera we are able to recover many points that are missed by the DSDT-SC, being able to process a higher number of frames. SD-DefSLAM assumes a global isometry by imposing a mesh, and in this case, as we do not update the mesh, it misses a big part of points when near-isometric deformation happens. Rigid methods focus in a small area of the scene being badly conditioned. In contrast, our direct method tracks almost the double of features in comparison with ORBSLAM.

The last one is Dataset 21 (from frame $\#$750) where the camera images two lobes of a liver moving as independent bodies, one lobe sliding over the other (See Fig.\ref{fig:niso}). Thanks to our formulation, the proposed deformable tracking can process global non-isometric deformations. We observe that our system is able to cope with deformations from independent bodies. In this case, SD-DefSLAM can track some of the points but with a high RMSE because its isometric deformation model cannot code the deformation actually observed. 

%% file: tables/tabledt.tex
\begin{table}[]
\caption{Comparison of our method against ORBSLAM and SD-DEFSLAM for the Hamlyn Dataset sequences 6, 20 and 21. We report RMSE and \# of frames processed.}
\begin{threeparttable}
\begin{tabular}{ll|r|r|r|r|r|}
\cline{3-7}
\multicolumn{2}{l|}{} & \multicolumn{2}{c|}{Rigid map}&
\multicolumn{3}{c|}{Deformable map}\\
\cline{3-7}
\multicolumn{2}{l|}{}                                          & ORB\cite{mur2017orb}  & DSDT-SM  & SD\cite{rodriguez2020sd} & DSDT-SC   & DSDT \\ \hline
\multicolumn{1}{|l|}{\multirow{2}{*}{6}}  & RMSE       & 4.85   &  3.26 & 2.72       & 9.24 & 3.17 \\ \cline{2-7} 
\multicolumn{1}{|l|}{}                            & \# Fr. & 128   &   200 & 286        & 334   & 300  \\ \hline \hline
\multicolumn{1}{|l|}{\multirow{2}{*}{20}} & RMSE       & 1.37   & 1.37  & 4.68       & 3.09  & 2.9 \\ \cline{2-7} 
\multicolumn{1}{|l|}{}                            & \# Fr. & 220   & 210    & 252        & 350   & 500  \\ \hline \hline
\multicolumn{1}{|l|}{\multirow{2}{*}{21}} & RMSE       & - & - & 6.19       & 1.81  & 1.30  \\ \cline{2-7} 
\multicolumn{1}{|l|}{}                            & \# Fr. & -   & - & 323        &321    & 300  \\ \hline
\end{tabular}
\begin{tablenotes}
        \item \textbf{ORB}: ORBSLAM; \textbf{DSDT-SM}: Direct Sparse with static map; \textbf{SD}: Semi-direct DefSLAM; \textbf{DSDT-SC}: Direct Sparse Deformable Tracking with static camera; \textbf{DSDT}: Direct Sparse Deformable Tracking 
\end{tablenotes}
\vspace{-0.5cm}
\end{threeparttable}
\end{table}

%% file: sections/conclusions.tex
\vspace{-0.15cm}
\section{Conclusion}
In this paper we have proposed a novel approach for  deformable tracking in deformable SLAM. Each map point is modeled as a 3D surfel that is a local approximation of the scene surface. The deformations of the map are modeled through the movement of these surfels. In contrast to the previous deformable tracking methods we have proposed to remove any connection between 3D map points.

We have proved experimentally that the local model for the deformable tracking can perform similarly to the state-of-the art methods and can perform more robustly and more accurately than the global methods in scenes composed of discontinuous surfaces, or with global non-isometric deformations. In addition, we reassert the potential of the direct methods over the feature-based equivalents.

Future work could extend this deformable tracking into a deformable mapping able to reconstruct scenes composed of discontinuous surfaces, or with global non-isometric deformations. With this two algorithmic components, it will be possible to create a new generation of monocular Deformable SLAM algorithms able to work in a wider range of scenarios.

\vspace{-0.15cm}